\documentclass{article}
\usepackage{spconf,amsmath,graphicx}
\usepackage{amsmath}
\usepackage{amssymb}
\usepackage{bbm}
\usepackage{xparse}
\usepackage{xspace}
\usepackage{url}
\usepackage{graphicx}
\usepackage{subfig}
\usepackage{verbatim}
\usepackage{booktabs}
\usepackage{tabularx}
\usepackage{todonotes}
\usepackage{enumitem}
\usepackage{multirow}
\usepackage{ragged2e}
\usepackage{colortbl}
\usepackage{textcomp}
\usepackage{subfig}

\usepackage{algorithm-mine}
\usepackage{algorithmic-mine}

\newcommand{\eqnname}{Eq.}
\newcommand{\sectionname}{Section}
\DeclareDocumentCommand\vec{m}{\boldsymbol{#1}}
\DeclareDocumentCommand\mat{m}{\boldsymbol{#1}}
\DeclareDocumentCommand\setvar{m}{\mathcal{#1}}
\DeclareDocumentCommand\t{m}{{#1}^{\rm T}}
\DeclareDocumentCommand\inv{m}{{#1}^{-1}}
\DeclareDocumentCommand\|{}{\mid} %{\,\vert\,}

\DeclareDocumentCommand\IndexedVar{m g g}{%
    #1\IfNoValueF{#2}{_{#2}}\IfNoValueF{#2}{^{(#3)}}
}
\DeclareDocumentCommand\Indicator{m}{\mathbbm{1}[#1]}

\DeclareDocumentCommand\E{o m}{\operatorname{E}\IfNoValueF{#1}{_#1}\left[#2\right]}
\DeclareDocumentCommand\N{o m m}{\mathcal{N} (\IfNoValueF{#1}{#1;} #2, #3)}

\DeclareDocumentCommand\MTState{o}{X\IfNoValueF{#1}{_{#1}}}
\DeclareDocumentCommand\MTMeasurement{o}{Z\IfNoValueF{#1}{_{#1}}}
\DeclareDocumentCommand\Clutter{o}{K\IfNoValueF{#1}{_{#1}}}
\DeclareDocumentCommand\NumTargetsSet{o}{M\IfNoValueF{#1}{_{#1}}}
\DeclareDocumentCommand\NumTargets{O{i}}{m_{#1}}
\DeclareDocumentCommand\STState{o}{\vec{x}\IfNoValueF{#1}{_{#1}}}
\DeclareDocumentCommand\STMeasurement{o}{\vec{z}\IfNoValueF{#1}{_{#1}}}

\DeclareDocumentCommand\Psurvival{o}{p_{\rm S}\IfNoValueF{#1}{(#1)}}
\DeclareDocumentCommand\Pdetection{o}{p_{\rm D}\IfNoValueF{#1}{(#1)}}
\DeclareDocumentCommand\STMotionModel{o}{f\IfNoValueF{#1}{_{#1}}}
\DeclareDocumentCommand\STMeasurementModel{o}{g\IfNoValueF{#1}{_{#1}}}
\DeclareDocumentCommand\STMotionProbability{o m m}{p\IfNoValueF{#1}{_{#1} (#2 \| #3)}}
\DeclareDocumentCommand\STMeasurementProbability{o m m}{p\IfNoValueF{#1}{_{#1}} (#2 \| #3)}

\DeclareDocumentCommand\v{o}{v\IfNoValueF{#1}{_{#1}}}       % state intensity
\DeclareDocumentCommand\b{o}{b\IfNoValueF{#1}{_{#1}}}       % birth intensity
\DeclareDocumentCommand\c{o}{c\IfNoValueF{#1}{_{#1}}}  % clutter intensity
\DeclareDocumentCommand\u{o}{u\IfNoValueF{#1}{_{#1}}}       % undetected targets

\DeclareDocumentCommand\Trajectory{o}{T\IfNoValueF{#1}{_{#1}}}
\DeclareDocumentCommand\TrackDistribution{o o}{\pi\IfNoValueF{#1}{_{#1}}\IfNoValueF{#2}{^{(#2)}}}
\DeclareDocumentCommand\TrackR{o o}{r\IfNoValueF{#1}{_{#1}}\IfNoValueF{#2}{^{(#2)}}}
\DeclareDocumentCommand\TrackP{o o}{p\IfNoValueF{#1}{_{#1}}\IfNoValueF{#2}{^{(#2)}}}
\DeclareDocumentCommand\TrackV{o o}{v\IfNoValueF{#1}{_{#1}}\IfNoValueF{#2}{^{(#2)}}}
\DeclareDocumentCommand\TrackS{o}{q\IfNoValueF{#1}{_{#1}}}

\DeclareDocumentCommand\ParticleSample{o o}{\vec{x}\IfNoValueF{#1}{_{#1}}\IfNoValueF{#2}{^{(#2)}}}
\DeclareDocumentCommand\ParticleWeightV{o o}{w\IfNoValueF{#1}{_{#1}}\IfNoValueF{#2}{^{(#2)}}}
\DeclareDocumentCommand\ParticleSet{o o}{Q\IfNoValueF{#1}{_{#1}}\IfNoValueF{#2}{^{(#2)}}}

\DeclareDocumentCommand\BirthSample{o o}{\vec{x}_{\textrm{b},\IfNoValueF{#1}{#1}}\IfNoValueF{#2}{^{(#2)}}}
\DeclareDocumentCommand\BirthWeightV{o o}{w_{\textrm{b},\IfNoValueF{#1}{#1}}^{(\IfNoValueF{#2}{#2})}}
\DeclareDocumentCommand\BirthSet{o}{B\IfNoValueF{#1}{_{#1}}}

\newcommand{\etal}{et al.}

\DeclareDocumentCommand\LossVar{}{\mathcal{L}}

\DeclareDocumentCommand{\Dataset}{}{\mathcal{D}}

\DeclareDocumentCommand\Input{}{\vec{x}}
\DeclareDocumentCommand\Label{}{y}
\DeclareDocumentCommand\Latent{}{\vec{r}}

\DeclareDocumentCommand\Encode{o m}{f(#2\IfNoValueF{#1}{; #1})}

\DeclareDocumentCommand\Predict{o o m}{g\IfNoValueF{#1}{_{#1}}(#3\IfNoValueF{#2}{; #2})}

\DeclareDocumentCommand\RunningLabel{}{k}
\DeclareDocumentCommand\NumLabels{}{L}
\DeclareDocumentCommand\NumSamples{}{M}

\DeclareDocumentCommand\ClassMean{}{\vec{\mu}}
\DeclareDocumentCommand\Stddev{}{\sigma}
\DeclareDocumentCommand\Variance{}{\Stddev^2}

\DeclareDocumentCommand\IdentityMatrix{}{\mat{I}}

\title{Simple Online and Realtime Tracking with a Deep Association Metric}
\name{%
    Nicolai Wojke\textsuperscript{\textdagger},
    Alex Bewley\textsuperscript{$\diamond$},
    Dietrich Paulus\textsuperscript{\textdagger}}
\address{%
    University of Koblenz-Landau\textsuperscript{\textdagger},
    Queensland University of Technology\textsuperscript{$\diamond$}}
\begin{document}
%\ninept
%
\maketitle
\begin{abstract}
Simple Online and Realtime Tracking (SORT) is a pragmatic approach to
multiple object tracking with a focus on simple, effective algorithms.
In this paper, we integrate appearance information to improve the performance
of SORT{}.
Due to this extension we are able to track objects through longer periods of
occlusions, effectively reducing the number of identity switches.
In spirit of the original framework we place much of the computational
complexity into an offline pre-training stage where we learn a deep
association metric on a large-scale person re-identification dataset.
During online application, we establish measurement-to-track associations
using nearest neighbor queries in visual appearance space.
Experimental evaluation shows that our extensions reduce the number of identity
switches by 45\%, achieving overall competitive performance at high
frame rates.
%Our evaluation on a large-scale pedestrian tracking dataset reveals improved
%performance at high frame rates.
\end{abstract}
\begin{keywords}
    Computer Vision,
    Multiple Object Tracking,
    %Convolutional Neural Networks,
    Data Association
\end{keywords}
\section{INTRODUCTION}
\label{sec:intro}

Due to recent progress in object detection, tracking-by-detection has become
the leading paradigm in multiple object tracking.
%Within this paradigm, object trajectories are usually
%optimized directly in image space, without additional state estimation.
Within this paradigm, object trajectories are usually found in a global
optimization problem that processes entire video batches at once.
For example, flow network formulations~\cite{%
ZhangLN08, PirsiavashRF11, DBLP:journals/pami/BerclazFTF11}
and probabilistic graphical models~\cite{%
DBLP:conf/cvpr/YangN12a, DBLP:conf/cvpr/YangN12,
DBLP:conf/cvpr/AndriyenkoSR12, DBLP:conf/cvpr/MilanSR13}
have become popular frameworks of this type.
%However, these methods process entire batches of video sequences and are not
%applicable in online scenarios where a target identity must be available
%at each time step.
However, due to batch processing, these methods are not applicable in
online scenarios where a target identity must be available at each time step.
More traditional methods are Multiple Hypothesis Tracking~(MHT)~\cite{%
Reid79analgorithm} and the Joint Probabilistic Data Association
Filter~(JPDAF)~\cite{fortmann1983sonar}. These methods perform data association
on a frame-by-frame basis.
In the JPDAF, a single state hypothesis is generated by weighting individual
measurements by their association likelihoods. In MHT, all possible
hypotheses are tracked, but pruning schemes must be applied for computational
tractability.
Both methods have recently been revisited in a tracking-by-detection
scenario~\cite{kim2015multiple, hamid2015joint} and shown promising results.
However, the performance of these methods comes at increased computational
and implementation complexity.

Simple online and realtime tracking (SORT)~\cite{Bewley2016_sort} is a much
simpler framework that performs Kalman filtering in image space and
frame-by-frame data association using the Hungarian method
with an association metric that measures bounding box overlap.
%intersection over union association metric.
This simple approach achieves favorable performance at high frame rates.
On the MOT challenge dataset~\cite{MOTChallenge2015},
SORT with a state-of-the-art people detector~\cite{renNIPS15fasterrcnn}
ranks on average higher than MHT on standard detections.
This not only underlines the influence of object detector performance on
overall tracking results, but is also an important insight from a practitioners
point of view.
%In a robotics context, similar observations have recently been made during an 
%extensive evaluation of different tracking
%algorithms~\cite{DBLP:conf/icra/LinderBLA16}. Their take-home message is that
%computational resources are better spent outside the actual tracking algorithms
%itself.

\begin{figure}[t!]
\centering
%\subfloat[][Frame 309]{
%    \includegraphics[width=0.45\linewidth]{images/occlusion1}
%}
%\quad
%\subfloat[][Frame 325]{
%    \includegraphics[width=0.45\linewidth]{images/occlusion2}
%}
%\caption{Example output of our tracker. Track ID is maintained through
%    several occlusions \dots}
\includegraphics[width=\linewidth]{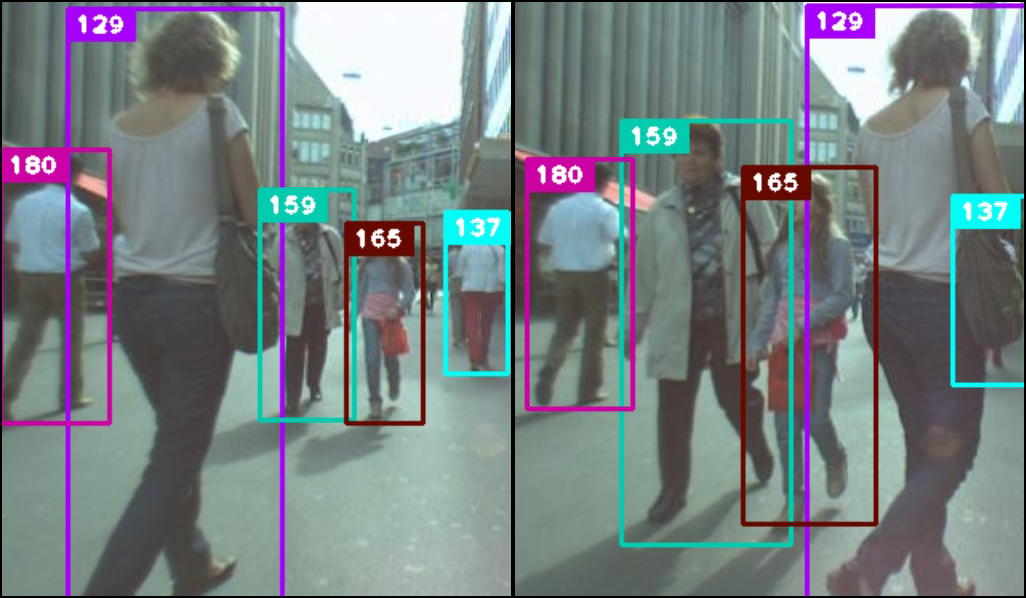}
\caption{Exemplary output of our method on the MOT challenge
    dataset~\cite{milan2016mot16} in a common tracking situation with
    frequent occlusion.}
\label{fig:example-output}
\vspace{-0.4cm}
\end{figure}

While achieving overall good performance in terms of tracking precision and
accuracy, SORT returns a relatively high number of identity switches.
This is, because the employed %intersection over union
association metric is only accurate when state estimation uncertainty is low.
Therefore, SORT has a deficiency in tracking through occlusions as they
typically appear in frontal-view camera scenes.
%In this paper, we exchange the association metric of SORT to incorporate
% appearance information to overcome this issue.
We overcome this issue by replacing the association metric with
a more informed metric that combines motion and appearance information.
In particular, we apply a convolutional neural network (CNN) that has been
trained to discriminate pedestrians on a large-scale person re-identification
dataset.
Through integration of this network we increase robustness against misses
and occlusions while keeping the system easy to implement, efficient, and
applicable to online scenarios.
Our code and a pre-trained CNN model are made publicly available to
facilitate research experimentation and practical application development.

\section{SORT WITH DEEP ASSOCIATION METRIC}
\label{sec:tracking}

We adopt a conventional single hypothesis tracking methodology with recursive
Kalman filtering and frame-by-frame data association.
In the following section we describe the core components of this system in
greater detail.

\subsection{Track Handling and State Estimation}

The track handling and Kalman filtering framework is mostly identical to the
original formulation in~\cite{Bewley2016_sort}.
We assume a very general tracking scenario where the camera is uncalibrated
and where we have no ego-motion information available.
While these circumstances pose a challenge to the filtering framework,
it is the most common setup considered in recent multiple object tracking
benchmarks~\cite{milan2016mot16}.
Therefore, our tracking scenario is defined on the eight dimensional
state space~$(u, v, \gamma, h, \dot{x}, \dot{y}, \dot{\gamma}, \dot{h})$
that contains the bounding box center position $(u, v)$, aspect ratio $\gamma$,
height $h$, and their respective velocities in image coordinates.
We use a standard Kalman filter with constant velocity motion and linear
observation model, where we take the bounding
coordinates~$(u, v, \gamma, h)$ as direct observations of the
object state.

For each track $k$ we count the number of frames since the last successful
measurement association $a_k$. This counter is incremented during Kalman filter
prediction and reset to~0 when the track has been associated with a measurement.
Tracks that exceed a predefined maximum age~$A_{\rm max}$ are considered to
have left the scene and are deleted from the track set.
New track hypotheses are initiated for each detection that cannot be associated
to an existing track.
These new tracks are classified as tentative during their first three frames.
During this time, we expect a successful measurement association at each time
step.  % to confirm the track.
Tracks that are not successfully associated to a measurement within their first
three frames are deleted.

\subsection{Assignment Problem}

A conventional way to solve the association between the
predicted Kalman states and newly arrived measurements is to build an
assignment problem that can be solved using the Hungarian algorithm.
Into this problem formulation we integrate motion and appearance information
through combination of two appropriate metrics.

To incorporate motion information we use the (squared) Mahalanobis distance
between predicted Kalman states and newly arrived measurements:
\begin{equation}
    d^{(1)}(i,j)
    =
    \t{(\vec{d}_j - \vec{y}_i)}
    \inv{\mat{S}_i}
    (\vec{d}_j - \vec{y}_i),
\end{equation}
where we denote the projection of the $i$-th track distribution into
measurement space by~$(\vec{y}_i, \mat{S}_i)$ and the $j$-th bounding box
detection by $\vec{d}_j$.
The Mahalanobis distance takes state estimation uncertainty into account by
measuring how many standard deviations the detection is away from the mean
track location.
Further, using this metric it is possible to exclude unlikely associations by
thresholding the Mahalanobis distance at a $95\%$ confidence interval computed
from the inverse $\chi^2$ distribution.
We denote this decision with an indicator
\begin{equation}
    b_{i,j}^{(1)} = \Indicator{d^{(1)}(i, j) \leq t^{(1)}}
\end{equation}
that evaluates to $1$ if the association between the $i$-th track and $j$-th
detection is admissible. For our four dimensional measurement space the
corresponding Mahalanobis threshold is~$t^{(1)} = 9.4877$.

While the Mahalanobis distance is a suitable association metric when motion
uncertainty is low, in our image-space problem formulation the predicted
state distribution obtained from the Kalman filtering framework provides only
a rough estimate of the object location.
In particular, unaccounted camera motion can introduce rapid displacements in
the image plane, making the Mahalanobis distance a rather uninformed metric for
tracking through occlusions.
Therefore, we integrate a second metric into the assignment problem.
For each bounding box detection $\vec{d}_j$ we compute an
appearance descriptor $\vec{r}_j$ with $\lVert\vec{r}_j\rVert = 1$.
Further, we keep a gallery~$\mathcal{R}_k = \{\vec{r}_k^{(i)}\}_{k=1}^{L_k}$
of the last $L_k=100$ associated appearance descriptors for each track $k$.
Then, our second metric measures the smallest cosine distance between
the~$i$-th track and~$j$-th detection in appearance space:
\begin{equation}
    d^{(2)}(i, j)
    =
    \min
    \{1 - \t{\vec{r}_j} \vec{r}^{(i)}_k \| \vec{r}^{(i)}_k\in \mathcal{R}_i \}.
    \label{eq:cosine-metric}
\end{equation}
Again, we introduce a binary variable to indicate if an association is admissible
according to this metric
\begin{equation}
    b_{i,j}^{(2)} = \Indicator{d^{(2)}(i, j) \leq t^{(2)}}
\end{equation}
and we find a suitable threshold for this indicator on a separate training
dataset.
In practice, we apply a pre-trained CNN to compute bounding box appearance
descriptors. The architecture of this network
is described in \sectionname~\ref{subsec:metric-learning}.

In combination, both metrics complement each other by serving different aspects
of the assignment problem. On the one hand, the Mahalanobis distance provides
information about possible object locations based on motion that
are particularly useful for short-term predictions.
%On the other hand, the cosine distance provides a more global compability score
%based on object appearance.
On the other hand, the cosine distance considers appearance information that
are particularly useful to recover identities after long-term occlusions, when
motion is less discriminative.
To build the association problem we combine both metrics using a weighted sum
\begin{equation}
    c_{i,j}
    =
    \lambda \, d^{(1)}(i, j) + (1 - \lambda) d^{(2)}(i, j)
    \label{eq:total_cost_elem}
\end{equation}
where we call an association admissible if it is within the gating region of
both metrics:
\begin{equation}
    b_{i,j} = \prod_{m=1}^{2} b_{i, j}^{(m)}.
    \label{eq:gate_elem}
\end{equation}
The influence of each metric on the combined association cost can be controlled
through hyperparameter $\lambda$.
During our experiments we found that setting $\lambda=0$ is a reasonable
choice when there is substantial camera motion.
In this setting, only appearance information are used in the association cost
term. However, the Mahalanobis gate is still used to disregarded infeasible
assignments based on possible object locations inferred by the Kalman filter.

\subsection{Matching Cascade}
\label{subseq:matching-cascade}

\begin{algorithm}[tb]
\caption{Matching Cascade}
\algsetup{linenosize=\small}
\small
\begin{algorithmic}[1]
    \INPUT{%
        Track indices $\setvar{T}=\{1, \dots, N\}$,
        Detection indices $\setvar{D}=\{1, \dots, M\}$,
        Maximum age~$A_{\rm max}$
    }
    \STATE{Compute cost matrix $\mat{C} = [c_{i,j}]$ using \eqnname~\ref{eq:total_cost_elem}} \label{lst:cascade_cost_matrix}
    \STATE{Compute gate matrix $\mat{B} = [b_{i,j}]$ using \eqnname~\ref{eq:gate_elem}} \label{lst:cascade_gate_matrix}
    \STATE{Initialize set of matches $\setvar{M} \gets \emptyset$}
    \STATE{Initialize set of unmatched detections $\setvar{U} \gets \setvar{D}$}
    \FOR{$n\in\{1,\dots,A_{\rm max}\}$}
        \STATE{Select tracks by age $\setvar{T}_n \gets \{i \in \setvar{T} \mid a_i = n\}$} \label{lst:cascade_select} \label{lst:cascade_track_selection}
        \STATE{$[x_{i,j}] \gets \operatorname{min\_cost\_matching}(\mat{C}, \setvar{T}_n, \setvar{U})$} \label{lst:cascade_solve}
        \STATE{$\mathcal{M} \gets \mathcal{M} \cup \{(i, j) \mid b_{i,j}\cdot x_{i,j} > 0 \}$} \label{lst:cascade_aggregate_matches}
        \STATE{$\mathcal{U} \gets \mathcal{U} \setminus \{j \mid \sum_i b_{i,j}\cdot x_{i,j} > 0\}$} \label{lst:cascade_update_unmatched}
    \ENDFOR{}
    \RETURN $\mathcal{M}, \mathcal{U}$ \label{lst:cascade_return}
\end{algorithmic}
\label{lst:matching_cascade}
\end{algorithm}

Instead of solving for measurement-to-track associations in a global
assignment problem, we introduce a cascade that solves a series of subproblems.
To motivate this approach, consider the following situation:
When an object is occluded for a longer period of time, subsequent Kalman
filter predictions increase the uncertainty associated with the object
location. Consequently, probability mass spreads out in state space and the
observation likelihood becomes less peaked.
Intuitively, the association metric should account for this spread of
probability mass by increasing the measurement-to-track distance.
Counterintuitively, when two tracks compete for the same detection, the
Mahalanobis distance favors larger uncertainty, because it effectively reduces
the distance in standard deviations of any detection towards the projected
track mean.
%While this could  be fixed by adding a normalization constant,
%we would like to encode similar behavior into our cosine appearance metric
%as well:
This is an undesired behavior as it can lead to increased track fragmentations
and unstable tracks.
Therefore, we introduce a matching cascade that gives priority to more
frequently seen objects to encode our notion of probability spread in
the association likelihood.
%Tracks that have been seen more recently should have priority in the assignment
%to reduce fragmentations and unstable tracks. Therefore, we propose a more
%general approach that enforces the desired behavior through a series of
%linear assignments.

Listing~\ref{lst:matching_cascade} outlines our matching algorithm.
As input we provide the set of track~$\setvar{T}$ and detection~$\setvar{D}$
indices as well as the maximum age $A_{\rm max}$.
In lines~\ref{lst:cascade_cost_matrix} and~\ref{lst:cascade_gate_matrix} we
compute the association cost matrix and the matrix of admissible associations.
We then iterate over track age $n$ to solve a linear assignment problem
for tracks of increasing age.
In line~\ref{lst:cascade_track_selection} we select the subset of
tracks~$\setvar{T}_n$ that have not been associated with a detection in the
last~$n$ frames.
In line~\ref{lst:cascade_solve} we solve the linear assignment between tracks
in~$\setvar{T}_n$ and unmatched detections~$\setvar{U}$.
In lines~\ref{lst:cascade_aggregate_matches}
and~\ref{lst:cascade_update_unmatched} we update the set of matches and
unmatched detections,
which we return after completion in line~\ref{lst:cascade_return}.
Note that this matching cascade gives priority to tracks of smaller age, i.e.,
tracks that have been seen more recently.
%, to encode our notion of probability
%spread in the association likelihood.

%In order to account for sudden appearance changes, e.g., due to partial
%occlusion with static scene geometry, we run a final stage of
%intersection over union~(IOU) association as proposed in the original SORT
%algorithm~\cite{Bewley2016_sort} on the set of unmatched detections and
%tracks of age $a_i=1$.
%We also exclude unconfirmed tracks from the matching cascade and only add them
%to the final IOU matching stage, to increase robustness against errorneous
%initialization.
In a final matching stage, we run intersection over union association
as proposed in the original SORT algorithm~\cite{Bewley2016_sort}
on the set of unconfirmed and unmatched tracks of age $n=1$. This helps to
to account for sudden appearance changes, e.g., due to partial occlusion with
static scene geometry, and to increase robustness against erroneous
initialization.

\subsection{Deep Appearance Descriptor}
\label{subsec:metric-learning}

\begin{table}[t!]
    \begin{tabular}{p{0.28\linewidth}cc}
        \toprule
        \textbf{Name} & \textbf{Patch Size/Stride} & \textbf{Output Size} \\
        \midrule
        Conv 1 & $3\times 3$/$1$ & $32\times 128\times 64$ \\
        Conv 2 & $3\times 3$/$1$ & $32\times 128\times 64$ \\
        Max Pool 3 & $3\times 3$/$2$ & $32\times 64\times 32$ \\
        Residual 4 & $3\times 3$/$1$ & $32\times 64\times 32$ \\
        Residual 5 & $3\times 3$/$1$ & $32\times 64\times 32$ \\
        Residual 6 & $3\times 3$/$2$ & $64\times 32\times 16$ \\
        Residual 7 & $3\times 3$/$1$ & $64\times 32\times 16$ \\
        Residual 8 & $3\times 3$/$2$ & $128\times 16\times 8$ \\
        Residual 9 & $3\times 3$/$1$ & $128\times 16\times 8$ \\
        Dense 10 & & $128$ \\
        \multicolumn{2}{l}{Batch and $\ell_2$ normalization} & $128$ \\
        \bottomrule
    \end{tabular}
    \caption{Overview of the CNN architecture. The final batch and $\ell_2$
    normalization projects features onto the unit hypersphere.}
\label{tab:network-architecture}
\end{table}

\begin{table*}[t!]
    \centering
    \begin{tabular}{@{}lccccccccc@{\hskip7pt}c@{}}
\toprule
 & \multicolumn{1}{l}{} & \textbf{MOTA} $\uparrow$ & \textbf{MOTP} $\uparrow$ & \textbf{MT} $\uparrow$ & \textbf{ML} $\downarrow$ & \textbf{ID} $\downarrow$ & \textbf{FM} $\downarrow$ & \textbf{FP} $\downarrow$ & \textbf{FN} $\downarrow$ & \textbf{Runtime} $\uparrow$ \\
\midrule
KDNT~\cite{yu2016poi}\textsuperscript{$\star$} & BATCH & 68.2 & 79.4 & 41.0\% & 19.0\% & 933 & 1093 & 11479 & 45605 & 0.7\,Hz \\
LMP\_p~\cite{keuper2016multi}\textsuperscript{$\star$} & BATCH & \textbf{71.0} & \textbf{80.2} & \textbf{46.9\%} & 21.9\% & 434 & \textbf{587} & 7880 & \textbf{44564} & 0.5\,Hz \\
MCMOT\_HDM~\cite{lee2016multi} & BATCH & 62.4 & 78.3 & 31.5\% & 24.2\% & 1394 & 1318 & 9855 & 57257 & 35\,Hz \\
NOMTwSDP16~\cite{choi2015near} & BATCH & 62.2 & 79.6 & 32.5\% & 31.1\% & \textbf{406} & 642 & \textbf{5119} & 63352 & 3\,Hz \\
\midrule
EAMTT~\cite{sanchez2016online} & \textbf{ONLINE} & 52.5 & 78.8 & 19.0\% & 34.9\% & 910 & \textbf{1321} & \textbf{4407} & 81223 & 12\,Hz \\
POI~\cite{yu2016poi}\textsuperscript{$\star$} & \textbf{ONLINE} & \textbf{66.1} & 79.5 & \textbf{34.0\%} & 20.8\% & 805 & 3093 & 5061 & \textbf{55914} & 10\,Hz \\
        SORT~\cite{Bewley2016_sort}\textsuperscript{$\star$} & \textbf{ONLINE} & 59.8 & \textbf{79.6} & 25.4\% & 22.7\% & 1423 & 1835 & 8698 & 63245 & \textbf{60\,Hz} \\
Deep SORT (Ours)\textsuperscript{$\star$} & \textbf{ONLINE} & 61.4 & 79.1 & 32.8\% & \textbf{18.2\%} & \textbf{781} & 2008 & 12852 & 56668 & 40\,Hz \\
\bottomrule
\end{tabular}
    \caption{Tracking results on the MOT16~\cite{milan2016mot16} challenge.
    We compare to other published methods with non-standard detections.
    The full table of results can be found on the challenge website.
    Methods marked with \textsuperscript{$\star$} use detections provided by~\cite{yu2016poi}.}
\label{tab:mot-results}
\end{table*}

By using simple nearest neighbor queries without additional metric learning,
successful application of our method requires a well-discriminating feature
embedding to be trained offline, before the actual online tracking application.
To this end, we employ a CNN that has been
trained on a large-scale person re-identification
dataset~\cite{Zheng2016} that contains over~1,100,000 images of~1,261
pedestrians, making it well suited for deep metric learning in a people
tracking context.

The CNN architecture of our network is shown in
\tablename~\ref{tab:network-architecture}.
In summary, we employ a wide residual network~\cite{Zagoruyko2016}
with two convolutional layers followed by six residual blocks. The global
feauture map of dimensionality $128$ is computed in dense layer 10.
A final batch and~$\ell_2$ normalization projects features
onto the unit hypersphere to be compatible with our cosine appearance
metric.
In total, the network has 2,800,864 parameters and one forward pass
of 32 bounding boxes takes approximately~$30\,\textrm{ms}$ on an Nvidia GeForce
GTX 1050 mobile GPU{}. Thus, this network is well suited for online tracking,
provided that a modern GPU is available.
%In total, the network has 2,800,864 parameters and one forward pass
%of 32 bounding boxes takes approximately~$15\,\textrm{ms}$ on an Nvidia Quadro
%M6000 GPU{}. Thus, this network is well suited for online tracking, provided
%that a modern GPU is available.
% NOTE(nwojke): On a GeForce GTX 1050 32 bounding boxes take
% approximately 30 ms.
%During training, a softmax classifier has been placed on top of the network to
%discriminate the identities contained in the dataset.
While the details of our training procedure are out of the scope of this paper,
we provide a pre-trained model in our GitHub
repository~\footnote{\url{https://github.com/nwojke/deep_sort}}
along with a script that can be used to generate features.

\section{EXPERIMENTS}
\label{sec:experiments}

We assess the performance of our tracker on the MOT16
benchmark~\cite{milan2016mot16}.
This benchmark evaluates tracking performance on seven challenging test
sequences, including frontal-view scenes with moving camera as well as top-down
surveillance setups.
As input to our tracker we rely on detections provided
by~Yu~\etal~\cite{yu2016poi}. They have trained a Faster RCNN on a collection
of public and private datasets to provide excellent performance.
For a fair comparison, we have re-run SORT on the same detections.

Evaluation on test sequences were carried out using~$\lambda=0$
and~$A_{\rm max}=30$ frames.
As in~\cite{yu2016poi}, detections have been thresholded at a confidence score
of~$0.3$.
The remaining parameters of our method have been found on separate training
sequences which are provided by the benchmark.
Evaluation is carried out according to the following metrics:
\begin{itemize}[leftmargin=*, noitemsep, topsep=0pt, parsep=0pt, partopsep=0pt]
\item Multi-object tracking accuracy (MOTA): Summary of overall tracking
    accuracy in terms of false positives, false negatives and identity
    switches~\cite{DBLP:journals/ejivp/BernardinS08}.
\item Multi-object tracking precision (MOTP): Summary of overall tracking
    precision in terms of bounding box overlap between ground-truth and
    reported location~\cite{DBLP:journals/ejivp/BernardinS08}.
\item Mostly tracked (MT): Percentage of ground-truth tracks that have the same
    label for at least 80\% of their life span.
\item Mostly lost(ML): Percentage of ground-truth tracks that are tracked for
    at most 20\% of their life span.
\item Identity switches (ID): Number of times the reported identity of a
    ground-truth track changes.
\item Fragmentation (FM): Number of times a track is interrupted by a missing
    detection.
\end{itemize}
The results of our evaluation are shown in \tablename~\ref{tab:mot-results}.
Our adaptions successfully reduce the number of identity switches.
In comparison to SORT, ID switches reduce from 1423 to 781.
This is a decrease of approximately 45\%.
At the same time, track fragmentation increase slightly due to maintaining
object identities through occlusions and misses.
We also see a significant increase in number of mostly tracked objects and
a decrease of mostly lost objects.
Overall, due to integration of appearance information we successfully maintain
identities through longer occlusions.
This can also be seen by qualitative analysis of the tracking output that
we provide in the supplementary material.
An exemplary output of our tracker is shown in
\figurename~\ref{fig:example-output}.

Our method is also a strong competitor to other online tracking frameworks.
In particular, our approach returns the fewest number of identity switches of
all online methods while maintaining competitive MOTA scores, track
fragmentations, and false negatives.
The reported tracking accuracy is mostly impaired by a larger number of false
positives.
Given their overall impact on the MOTA score, applying a larger confidence
threshold to the detections can potentially increase the reported performance
of our algorithm by a large margin.
However, visual inspection of the tracking output shows that these false
positives are mostly generated from sporadic detector responses at static scene
geometry.
Due to our relatively large
maximum allowed track age, these are more commonly joined to object
trajectories. At the same time, we did not observe tracks jumping between
false alarms frequently. Instead, the tracker commonly generated relatively
stable, stationary tracks at the reported object location.

%\subsection{Runtime}

% Measured on MOT16-02:
% * 600 frames
% * Total time tracker: 5.75 s
% * Total num detections (confidence >= 0.8): 4412 detections
% * Every 32 detections take approx. 15 ms -> +2.068125s total time
% * (5.75 + 2.068125) / 600 = 0.013 s per frame = approx. 76.7 Hz

% MOT16:
% On average (24.8 detections per frame / 32 ms) * 15 ms -> 11.625 ms CNN.
% The rest runs at 77.7 fps -> 12.87 ms.
% Total ms: 12.9 + 16.3 = 24.49 ms = 40.825 fps
Our implementation runs at approximately 20\,Hz with roughly half of the time
spent on feature generation.
Therefore, given a modern GPU, the system remains computationally efficient
and operates at real time.

\section{CONCLUSION}
\label{sec:conclusion}

We have presented an extension to SORT that incorporates appearance information
through a pre-trained association metric.
Due to this extension, we are able to track through longer periods of
occlusion, making SORT a strong competitor to state-of-the-art online
tracking algorithms. Yet, the algorithm remains simple to implement and
runs in real time.

\clearpage

\bibliographystyle{IEEEbib}
\bibliography{root}

\end{document}